\documentclass{article} 
\usepackage{iclr2027_conference,times}

\usepackage{amsmath,amsfonts,bm}

\def\eqref#1{equation~\ref{#1}}

\def\1{\bm{1}}

\DeclareMathAlphabet{\mathsfit}{\encodingdefault}{\sfdefault}{m}{sl}
\SetMathAlphabet{\mathsfit}{bold}{\encodingdefault}{\sfdefault}{bx}{n}

\usepackage{hyperref}
\usepackage{graphicx}
\usepackage{url}
\usepackage{float}
\usepackage{booktabs}
\usepackage{multirow}  
\usepackage{array}      
\usepackage{subcaption}
\usepackage{makecell}  
\usepackage{comment}  
\usepackage{amsmath,amssymb,amsthm}

\theoremstyle{definition}

\theoremstyle{remark}

\usepackage[table]{xcolor} 
\usepackage{xcolor}
\usepackage{wrapfig}

\title{Rethinking Soft Tokens for Parallel \\ Decoding in Diffusion Language Models}

\author{
Kodai Kawamura \quad
Kenji Kawaguchi \quad
Anji Liu \\
National University of Singapore
}

\iclrfinalcopy 
\begin{document}

\maketitle

\begin{abstract} 
Diffusion language models (DLMs) enable parallel generation by predicting and committing multiple tokens at each denoising step, yet they can generate individually plausible but mutually inconsistent tokens.
Recent work shows that \emph{soft tokens} can mitigate this issue by representing uncertain positions with continuous embeddings built from the model's predictive distribution at the previous decoding step.
However, although soft tokens are commonly understood as preserving predictive uncertainty, how soft-token feedback improves parallel decoding has not been systematically examined. 
In this paper, we investigate this question in frozen pretrained DLMs to examine soft-token feedback without the effects of additional training.
To construct soft-token inputs in a training-free setting, we identify a geometric mismatch between conventional soft-token construction and the pretrained embedding space. 
Based on this observation, we propose a training-free, geometry-aware construction of soft tokens.
Our analysis of soft-token feedback suggests that uncertainty preservation alone does not fully explain how it reshapes subsequent predictions.
To better explain how soft-token feedback improves parallel decoding, we provide empirical evidence that it favors coherent token sequences.
Across four pretrained DLMs and four math and code benchmarks, our method outperforms standard parallel decoding and a training-free Euclidean soft-token baseline. 
Code: \url{https://github.com/kodaikawamura/rethinking-soft-tokens}

\end{abstract}

\section{Introduction}
Diffusion language models (DLMs) have recently emerged as a promising alternative to autoregressive language models by enabling parallel text generation through iterative denoising~\citep{nie2025large,ye2025dream}. 
By predicting multiple masked positions simultaneously, DLMs naturally support parallel decoding and offer the potential for substantially faster decoding~\citep{wu2025fast,benhamu2025accelerated}.
However, standard parallel decoding uses a factorized approximation to the joint distribution, ignoring dependencies among simultaneously predicted tokens~\citep{israel2025accelerating}. 
As a result, DLMs can commit multiple tokens that are individually plausible but mutually inconsistent, introducing errors that can propagate through later denoising steps~\citep{liu2025discrete}. 

To mitigate this limitation, recent works introduce \emph{soft tokens}~\citep{hersche2026soft,zhong2026beyond,chen2026dmax}.
Rather than either replacing a masked position with a discrete prediction or leaving it unresolved as \texttt{[MASK]}, soft-token methods represent unresolved positions with continuous embeddings constructed by interpolating between the \texttt{[MASK]} embedding and a probability-weighted average of the top-$k$ predicted-token embeddings. 
These continuous representations are fed back into the model at the next denoising step and have been shown to improve parallel decoding. 

Despite their empirical success, a fundamental question remains: \emph{how does soft-token feedback improve parallel decoding?}
Although its benefits are commonly attributed to preserving predictive uncertainty~\citep{hersche2026soft,zhong2026beyond}, this explanation has not been systematically examined and does not specify how soft tokens affect subsequent predictions.
Existing methods train or fine-tune models to process soft-token representations, making it difficult to determine whether their gains arise from soft-token feedback itself or from learned adaptation to these representations.
To investigate how soft-token feedback reshapes predictions without the effects of additional training, we develop a soft-token construction that can be applied directly to frozen pretrained DLMs. 

To enable this investigation, we first address a question: \emph{how should soft tokens be constructed for models that have not been trained to process them?}
We find that directly applying conventional Euclidean soft-token constructions to frozen pretrained DLMs can distort angular information and shrink embedding norms.
We therefore propose geometry-aware soft tokens based on spherical interpolation, which can be directly incorporated into pretrained DLMs without additional training.
Across four pretrained DLMs and four math and code benchmarks, our method outperforms both standard parallel decoding and training-free Euclidean soft-token baselines. 

Using this training-free construction, we then investigate how soft-token feedback affects subsequent predictions.
In controlled experiments, we show that uncertainty preservation alone does not fully explain the observed behavior.
We further provide empirical evidence that soft-token feedback shifts probability mass toward coherent token sequences and away from inconsistent combinations. 
These findings offer a possible explanation for how soft-token feedback improves parallel decoding.

\section{Preliminaries}

\subsection{Diffusion Language Models}
\label{preliminary_DLM}
Let $\mathbf{x}_0=(x_0^1,\ldots,x_0^N)\in\mathcal{V}^N$ denote a clean sequence of tokens, where $\mathcal{V}$ is a vocabulary of size $V$ that includes a special \texttt{[MASK]} token.
In this paper, we focus on masked diffusion language models~\citep{sahoo2024simple}, in which the \texttt{[MASK]} token serves as an absorbing state during the forward diffusion process.

The model learns the data distribution $p_{\mathrm{data}}(\mathbf{x}_0)$ by recovering clean sequences from partially masked observations.
The forward diffusion process gradually corrupts a clean sequence by replacing its tokens with \texttt{[MASK]}:
\[
\mathbf{x}_0,\mathbf{x}_1,\ldots,\mathbf{x}_T,
\]
where $\mathbf{x}_t=(x_t^1,\ldots,x_t^N)\in\mathcal{V}^N$ represents the token sequence at diffusion step $t$.
The transition between two consecutive states is modeled as
\[
q(x_t^i \mid x_{t-1}^i)
=
Q_t[x_{t-1}^i,x_t^i],
\qquad
q(\mathbf{x}_t \mid \mathbf{x}_{t-1})
=
\prod_{i=1}^{N}
q(x_t^i \mid x_{t-1}^i),
\]
where $Q_t\in[0,1]^{V\times V}$ is a categorical transition matrix whose entry $Q_t[a,b]$ gives the probability of transitioning from token $a$ to token $b$.
In DLMs, $Q_t$ is constructed such that each token either remains unchanged or transitions to the absorbing \texttt{[MASK]} state, with the probability of being masked increasing over time. 

During inference, decoding begins from $\mathbf{x}_T$, where the prompt tokens remain fixed and all positions to be generated are initialized as \texttt{[MASK]}.
At each denoising step, the model maps each input token, including \texttt{[MASK]}, to its learned embedding and feeds the embedding sequence into a bidirectional Transformer to obtain a predictive distribution over the vocabulary at each position.
A common unmasking strategy selects the positions with the highest prediction confidence and replaces them with their predicted tokens, while the remaining positions stay masked for subsequent iterations.
This procedure supports parallel decoding by unmasking multiple positions in a denoising step.

\subsection{Challenges in Parallel Decoding}
Although parallel decoding is possible with diffusion language models, they can generate token predictions that are individually plausible yet mutually inconsistent.
Given a partially masked context $\mathbf{x}$, let $\mathbf{y}=(y_1,\ldots,y_m)$ denote the tokens predicted simultaneously.
Standard parallel decoding uses the factorized distribution $q_{\mathrm{parallel}}$:
\[
q_{\mathrm{parallel}}(\mathbf{y}\mid\mathbf{x})
=
\prod_{i=1}^{m}
p_\theta(y_i\mid\mathbf{x}),
\]
where $p_\theta$ is the predictive distribution of the model with parameters $\theta$.
This factorization ignores dependencies among simultaneously predicted tokens.
Even when every predicted marginal is exact, their product generally differs from the true joint distribution when the tokens are dependent.
Thus, fully factorized prediction introduces an approximation error that cannot be eliminated by improving the individual marginals alone~\citep{liu2025discrete,li2026breaking}.

\begin{figure*}[t]
    \centering
    \textbf{Prompt:} Answer is either Harry Potter or Star Wars:
    \texttt{[MASK]} \texttt{[MASK]}
    \par\smallskip
    \includegraphics[width=0.98\textwidth]{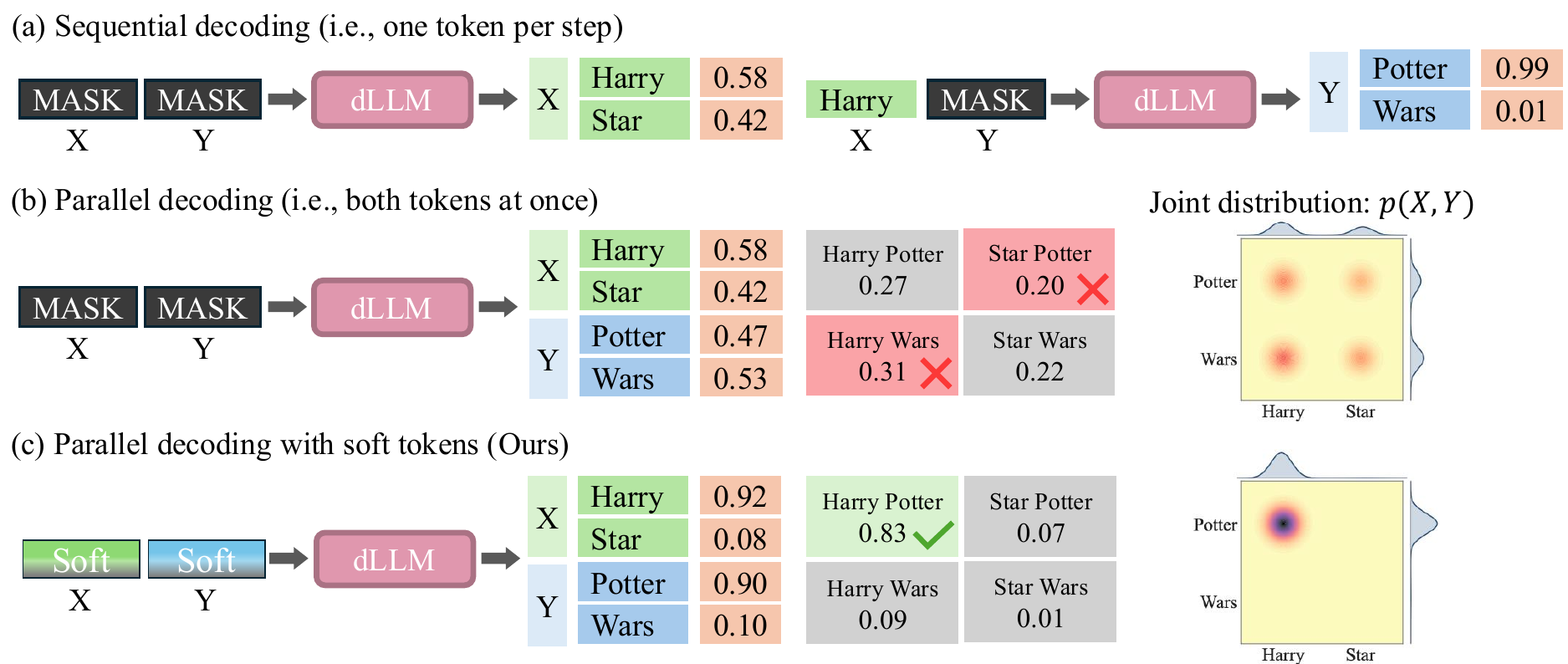}
    \caption{
    \textbf{Soft-token feedback improves sequence-level consistency.}
    (a) Sequential decoding conditions each prediction on previously committed tokens, producing the coherent sequence \emph{Harry Potter}. 
    (b) Parallel decoding predicts both tokens independently, assigning probability mass to inconsistent combinations and making \emph{Harry Wars} the most likely sequence.
    (c) Soft-token feedback using our proposed construction (Section~\ref{sec:method}) shifts probability mass toward the coherent sequence \emph{Harry Potter} and away from inconsistent combinations.
    }
    \label{fig:example}
    \vspace{-8pt}
\end{figure*}

\subsection{Soft-Token Decoding}
\label{preliminary-soft}
Recent work introduces \emph{soft tokens} and shows improved performance in parallel decoding, suggesting a potential way to mitigate the inconsistencies arising from independent token predictions~\citep{hersche2026soft,zhong2026beyond,chen2026dmax}.
Instead of retaining the discrete \texttt{[MASK]} token, soft-token decoding constructs a continuous representation from the model's predictive distribution and feeds it into the next denoising iteration.
This continuous representation retains information from multiple candidate tokens without committing to a single prediction. 

A conventional Euclidean soft token is constructed as~\citep{hersche2026soft,zhong2026beyond}
\[
\mathbf{e}_{\mathrm{soft}}
=
(1-\lambda)\mathbf{e}_{\mathrm{MASK}}
+
\lambda\sum_{i=1}^{k}p_i\mathbf{e}_i,
\]
where $\mathbf{e}_i$ is the embedding of the $i$-th token among the model's top-$k$ predicted tokens, $p_i$ is its predictive probability renormalized over these candidate tokens, and $\lambda\in[0,1]$ controls the interpolation with the \texttt{[MASK]} embedding $\mathbf{e}_{\mathrm{MASK}}$.
The resulting soft token replaces the \texttt{[MASK]} embedding in the next denoising iteration.
Existing methods either train DLMs from scratch with soft-token inputs or adapt pretrained DLMs to these inputs through additional training.

Figure~\ref{fig:example} illustrates how soft-token feedback, implemented in frozen DLMs using our proposed construction (Section~\ref{sec:method}), can mitigate errors in parallel decoding. 
We consider a toy experiment in which a prompt admits two coherent sequences: \emph{Harry Potter} and \emph{Star Wars}. 
Sequential decoding allows a DLM to account for the dependency between the two tokens by conditioning the second prediction on the first, but it is slow as the model commits one token per step. 
When both tokens are predicted independently in parallel, however, the product of their marginals assigns probability mass to inconsistent combinations such as \emph{Harry Wars} and \emph{Star Potter}, even when the marginal distribution at each position is exact. 
Soft-token feedback reshapes these marginals to favor mutually compatible token choices, making \emph{Harry Potter} the most likely sequence under parallel decoding.

These observations raise a question: \emph{how does soft-token feedback reshape subsequent predictions to improve sequence-level consistency?}
Although prior work attributes performance gains from soft-token feedback to uncertainty preservation~\citep{hersche2026soft,zhong2026beyond}, this explanation has not been systematically examined.
We first formalize this prevailing interpretation to provide a reference for examining this behavior.

\section{An Uncertainty-Preservation View of Soft Tokens}
\label{sec:preserve_uncertainty}
Soft tokens are commonly understood to preserve uncertainty by carrying information about multiple candidates into subsequent predictions.
To formalize this interpretation, we model a soft token as \emph{soft evidence}~\citep{chan2005revision,munk2023uncertain} over a hypothetical discrete input state
\[
Z\in\{\mathrm{M},v_1,\ldots,v_k\},
\]
where \(\mathrm{M}\) denotes the unresolved \texttt{[MASK]} state and \(v_i\) is the candidate token with the \(i\)-th highest predicted probability at the position represented by the soft token.

If a soft token preserves uncertainty in this probabilistic sense, meaning that its role is to represent uncertainty over the values of \(Z\), prediction marginalizes over these realizations:
\[
p(Y=y\mid E)
=
\sum_z
p(Z=z\mid E)\,
p_\theta(Y=y\mid E,Z=z),
\]
where \(E\) is the current decoding context.
For interpolation strength \(\lambda\) and candidate probabilities \(p_i\) renormalized over the top-\(k\) tokens, the distribution over \(Z\) is defined by 
\[
p(Z=v_i\mid E)=\lambda p_i,
\qquad
p(Z=\mathrm{M}\mid E)
=
1-\lambda\sum_{i=1}^{k}p_i.
\]
Therefore,
\[
p(Y=y\mid E)
=
\left(
1-\lambda\sum_{i=1}^{k}p_i
\right)
p_\theta(Y=y\mid E,Z=\mathrm{M})
+
\lambda\sum_{i=1}^{k}
p_i\,p_\theta(Y=y\mid E,Z=v_i).
\]

The same argument extends to multiple soft-token positions.
Let \(\mathcal{S}\) be the set of positions represented by soft tokens, and let \(\mathbf{Z}=(Z_j)_{j\in\mathcal{S}}\) denote their joint discrete realization.
Assuming that the positionwise evidence is realized independently,
\[
p(\mathbf{Z}=\mathbf{z}\mid E)
=
\prod_{j\in\mathcal{S}}
p(Z_j=z_j\mid E).
\]
Marginalizing over all joint discrete realizations yields an additive mixture of the predictive distributions they induce, which we denote by \(p_{\mathrm{add}}\):
\[
p_{\mathrm{add}}(y)
:=
p(Y=y\mid E)
=
\sum_{\mathbf{z}}
p(\mathbf{Z}=\mathbf{z}\mid E)
p_\theta(Y=y\mid E,\mathbf{Z}=\mathbf{z}).
\]
This additive mixture follows naturally from the probabilistic interpretation above.

We examine whether predictions induced by continuous soft-token inputs agree with this reference. 
To investigate this question without the effects of additional training, we study soft-token feedback in frozen pretrained DLMs.
This requires addressing how soft tokens should be constructed for models that have not been trained to process these representations.
We develop such a construction in the following section, then return to the question of how its feedback reshapes subsequent predictions.

\section{Training-Free Soft-Token Decoding}
\label{sec:method}
To enable this investigation, we address how to construct soft tokens for a frozen pretrained DLM without additional training.
Because the model has not been trained to process mixtures of token embeddings, the placement of these representations within the pretrained embedding space is particularly important.
We first identify the geometric mismatch introduced by conventional Euclidean construction and then propose a geometry-aware method for constructing soft tokens.

\subsection{Geometric Mismatch in Euclidean Soft Tokens}
\label{subsec:mismatch}
As discussed in Section~\ref{preliminary-soft}, conventional soft-token construction consists of two operations: aggregating the top-$k$ candidate embeddings and interpolating between the aggregated representation and the \texttt{[MASK]} embedding. 
To investigate the geometry relevant to these operations, we consider how the resulting input is processed by the frozen model. 
In all DLMs considered in this work, every self-attention and feed-forward sublayer operates on an RMS-normalized representation:
\[
\operatorname{RMSNorm}(\mathbf{h})
=
\left(\sqrt{d}\,\boldsymbol{\gamma}\right)
\odot
\frac{\mathbf{h}}{\lVert\mathbf{h}\rVert_2}.
\]
where $d$ is the hidden dimension, $\boldsymbol{\gamma}$ is a learned scaling vector, and $\odot$ denotes elementwise multiplication~\citep{zhang2019root}.
Through this operation, the input representation $\mathbf{h}$ is normalized by its magnitude before each Transformer sublayer, reducing sensitivity to the input norm while retaining directional information.

This repeated normalization motivates explicitly controlling the direction of soft-token representations.
Our empirical analysis reveals a geometric distinction between the two operations in conventional soft-token construction: top-$k$ candidate embeddings are relatively aligned with one another, whereas their aggregate is nearly orthogonal to the \texttt{[MASK]} embedding across the evaluated models and datasets (Section~\ref{sec:geometry_analysis}).

The relative alignment among candidate embeddings supports retaining Euclidean averaging as a practical choice for top-$k$ aggregation.
This choice is also supported by prior work demonstrating the effectiveness of probability-weighted embedding aggregation in frozen autoregressive language models without additional training~\citep{zhang2025soft}.

Interpolation with \texttt{[MASK]}, however, must bridge a much larger angular separation.
Under linear interpolation, the resulting direction depends on both the interpolation weight and the relative norms of the two endpoints.
Consequently, the same weight can produce different angular movements toward the candidate aggregate, while the embedding norm may also shrink (Appendix~\ref{sec:mask-interpolation-geometry}).
These observations motivate controlling direction and magnitude separately when interpolating between the candidate aggregate and the \texttt{[MASK]} embedding.

\subsection{Geometry-Aware Soft-Token Construction}
\label{subsec:construction}
We construct geometry-aware soft tokens in two stages: candidate aggregation and interpolation with the \texttt{[MASK]} embedding.
Following the analysis in Section~\ref{subsec:mismatch}, we retain Euclidean top-$k$ aggregation and compute the semantic representation as
\[
\mathbf{m}
=
\sum_{i=1}^{k}
p_i \mathbf{e}_i,
\]
where $p_i$ denotes the probability assigned to the $i$-th top-$k$ candidate token, whose embedding is $\mathbf{e}_i$.

We then interpolate between the \texttt{[MASK]} embedding and the semantic representation by explicitly controlling direction and norm separately. 
We first interpolate between their directions, then scale the resulting unit vector to match the \texttt{[MASK]} embedding norm, retaining the pretrained scale associated with an unresolved position.

For directional interpolation, we use spherical linear interpolation (SLERP)~\citep{shoemake1985animating}. Specifically, we normalize the two endpoints and compute their angular separation:
\[
\widehat{\mathbf{e}}_{\mathrm{MASK}}
=
\frac{\mathbf{e}_{\mathrm{MASK}}}
{\lVert\mathbf{e}_{\mathrm{MASK}}\rVert_2},
\qquad
\widehat{\mathbf{m}}
=
\frac{\mathbf{m}}
{\lVert\mathbf{m}\rVert_2},
\qquad
\theta
=
\arccos\left(
\widehat{\mathbf{e}}_{\mathrm{MASK}}^{\top}
\widehat{\mathbf{m}}
\right).
\]
The resulting soft-token embedding is given by
\[
\mathbf{e}_{\mathrm{soft}}
=
\lVert\mathbf{e}_{\mathrm{MASK}}\rVert_2
\left[
\frac{\sin((1-\lambda)\theta)}{\sin\theta}
\widehat{\mathbf{e}}_{\mathrm{MASK}}
+
\frac{\sin(\lambda\theta)}{\sin\theta}
\widehat{\mathbf{m}}
\right],
\qquad
\lambda\in[0,1].
\]
Here, $\lambda$ controls how far the soft token moves from the \texttt{[MASK]} direction toward the direction of the semantic representation along the spherical arc, while its norm remains fixed at $\lVert\mathbf{e}_{\mathrm{MASK}}\rVert_2$.

We use the proposed construction for soft tokens while keeping all pretrained model parameters frozen.
At each iteration, we commit discrete tokens at the positions with the highest prediction confidence, while each unresolved position is represented by a soft token recomputed from its updated top-$k$ predictive distribution. 

\section{Understanding Soft-Token Feedback}
\label{sec:mechanism}
Having introduced a training-free construction, we now return to the uncertainty-preservation interpretation in Section~\ref{sec:preserve_uncertainty}.
We first examine whether predictions induced by our construction follow the additive mixture implied by this interpretation.
We then investigate how the observed behavior relates to sequence-level consistency in parallel decoding.

\subsection{Limitations of the Uncertainty-Preservation Explanation}
\label{sec:mixture_behavior}
Representing multiple candidate tokens in a continuous input does not by itself determine how their information is combined in subsequent predictions.
Under the uncertainty-preservation interpretation in Section~\ref{sec:preserve_uncertainty}, this information is propagated by marginalizing over possible discrete input realizations, yielding the additive mixture $p_{\mathrm{add}}$.
To examine whether our soft-token construction follows this interpretation, we compare its predictions with the additive mixture $p_{\mathrm{add}}$ and the following normalized multiplicative mixture:
\[
p_{\mathrm{mult}}(y)
=
\frac{
\exp\!\left(
\sum_{\mathbf{z}}
p(\mathbf{Z}=\mathbf{z}\mid E)
\log p_\theta(Y=y\mid E,\mathbf{Z}=\mathbf{z})
\right)
}{
\displaystyle
\sum_{y'}
\exp\!\left(
\sum_{\mathbf{z}}
p(\mathbf{Z}=\mathbf{z}\mid E)
\log p_\theta(Y=y'\mid E,\mathbf{Z}=\mathbf{z})
\right)
}.
\]
\begin{wraptable}{r}{0.5\textwidth}
\vspace{-8pt}
\centering
\caption{
JS divergence between predictions from our construction and the reference mixtures.
Lower values indicate closer agreement; bold indicates the closer reference.
}
\label{tab:additive_multiplicative_js}
\small
\setlength{\tabcolsep}{6pt}
\begin{tabular}{@{}lcc@{}}
\toprule
Model & Additive & Multiplicative \\
\midrule
LLaDA
& 0.2099 & \textbf{0.0410} \\
LLaDA-1.5
& 0.0496 & \textbf{0.0019} \\
LLaDA-2.0 mini
& 0.2198 & \textbf{0.0835} \\
Dream
& 0.0566 & \textbf{0.0119} \\
\bottomrule
\end{tabular}
\end{wraptable}
We compare soft-token predictions with these reference mixtures using Jensen--Shannon (JS) divergence on controlled prompts (Section~\ref{sec:mechanism_analysis}).
Table~\ref{tab:additive_multiplicative_js} shows that predictions induced by our construction are closer to the multiplicative reference than to the additive mixture across all four models.
These results suggest that the uncertainty-preservation interpretation alone does not fully account for the predictive behavior induced by our construction.
In the following subsection, we examine how this multiplicative behavior may contribute to improved parallel decoding.

\subsection{Agreement-Seeking Feedback and Sequence-Level Consistency}
\label{sec:additive_multiplicative}

Multiplicative aggregation combines predictive distributions through a weighted average of their log probabilities.
Predictions assigned very low probability by input realizations with substantial weight can therefore be suppressed, even when other realizations support them.
This sensitivity to disagreement across input realizations motivates an \emph{agreement-seeking} explanation for how soft-token feedback improves parallel decoding.

More specifically, soft-token feedback allows each position to incorporate candidate information from other positions before commitment.
We hypothesize that its sensitivity to conflicting candidate evidence helps shift predictions across positions toward mutually compatible choices.
Although the resulting joint prediction remains factorized, soft-token feedback can make inconsistent cross-combinations less likely by favoring a coherent combination of tokens across positions.

To examine this sequence-level behavior, we construct a sequential reference from the same frozen model by averaging its joint predictions over all unmasking orders.
Each prediction conditions on previously filled tokens, allowing the reference to incorporate dependencies among answer tokens.
We evaluate agreement using conditional reverse KL (Appendix~\ref{app:toy_consistency}), which penalizes probability assigned to combinations poorly supported by the reference.
This criterion is relevant to parallel decoding because mode-seeking predictions can yield lower reverse KL by concentrating probability on a coherent alternative and assigning less probability to inconsistent cross-combinations.

The analysis in Section~\ref{sec:mechanism_analysis} shows that our method achieves lower reverse KL than vanilla parallel decoding and Euclidean soft-token baselines, indicating closer agreement with the sequential reference.
Together with the mixture analysis, these results support an \emph{agreement-seeking} interpretation: feedback shifts the marginals toward mutually compatible predictions before commitment.

\section{Experiments}
\label{sec:experiments}

\subsection{Settings}
\label{sec:setup}

\paragraph{Models.}
We evaluate four instruction-tuned diffusion language models: LLaDA-Instruct-8B~\citep{nie2025large}, LLaDA-1.5-Instruct~\citep{zhu2026llada}, LLaDA-2.0-mini~\citep{bie2025llada}, and Dream-7B~\citep{ye2025dream}.
All model parameters remain frozen throughout decoding.

\paragraph{Benchmarks.}
We conduct experiments on four mathematical-reasoning and code-generation benchmarks.
GSM8K~\citep{cobbe2021training} contains grade-school mathematical reasoning problems, while MATH500~\citep{lightman2024lets} contains competition-level mathematics problems.
HumanEval~\citep{chen2021evaluating} and MBPP~\citep{austin2021program} evaluate Python code generation from natural-language specifications.
We report exact-match accuracy for GSM8K and MATH500 and pass@$1$ for HumanEval and MBPP.

\paragraph{Baselines.}
We compare our method with vanilla parallel decoding, which commits predictions at the most-confident unresolved positions in each iteration while leaving the remaining positions as \texttt{[MASK]} tokens.
To the best of our knowledge, no existing soft-token decoding method can be directly applied to a frozen pretrained DLM without additional training.
We therefore construct a training-free Euclidean soft-token baseline. 
This baseline linearly interpolates between the \texttt{[MASK]} embedding and the aggregated candidate representation, whereas our manifold-aware method uses spherical interpolation. 
Additional implementation details are provided in Appendix~\ref{app:settings}.

\begin{table*}[t]
\centering
\caption{
Comparison of accuracy (\%) with baseline methods across four diffusion language models under parallel decoding settings.
Vanilla denotes standard parallel decoding without soft-token feedback, and Euclidean denotes the training-free soft-token baseline using linear interpolation in embedding space.
Tokens/step denotes the number of tokens committed per denoising iteration.
The best results for each model, benchmark, and decoding setting are shown in bold.
}
\vspace{-8pt}
\label{tab:main_accuracy}
\setlength{\tabcolsep}{3.5pt}
\resizebox{\textwidth}{!}{
\begin{tabular}{llcccccccccccc}
\toprule
\multirow{2}{*}[-0.5ex]{Model}
& Method
& \multicolumn{3}{c}{GSM8K}
& \multicolumn{3}{c}{MATH500}
& \multicolumn{3}{c}{HumanEval}
& \multicolumn{3}{c}{MBPP} \\
\cmidrule(lr){2-2}
\cmidrule(lr){3-5}
\cmidrule(lr){6-8}
\cmidrule(lr){9-11}
\cmidrule(lr){12-14}
& Tokens/step
& 2 & 4 & 8
& 2 & 4 & 8
& 2 & 4 & 8
& 2 & 4 & 8 \\
\midrule

\multirow{3}{*}{LLaDA-2.0 mini}
& Vanilla 
& 89.61 & 83.09 & 49.81
& 40.60 & 37.40 & 16.20
& 70.73 & 46.95 & \textbf{24.39}
& 55.80 & 42.80 & \textbf{23.80} \\
& Euclidean
& 89.76 & 85.82 & 51.78
& 43.60 & 37.20 & 18.40
& 69.51 & 50.00 & 22.56
& 57.60 & 43.80 & 23.00 \\
\rowcolor{gray!15}
\cellcolor{white}
& Ours
& \textbf{90.60} & \textbf{87.40} & \textbf{54.20}
& \textbf{44.60} & \textbf{39.00} & \textbf{18.80}
& \textbf{71.95} & \textbf{53.05} & 23.17
& \textbf{57.80} & \textbf{44.20} & 23.40 \\

\midrule

\multirow{3}{*}{LLaDA-1.5}
& Vanilla 
& \textbf{72.25} & 65.13 & 43.82
& 20.80 & 17.60 & \textbf{15.00}
& 38.41 & 24.39 & 12.80
& 34.80 & 26.20 & 17.20 \\
& Euclidean
& 71.40 & 66.79 & 47.54
& 21.00 & 18.80 & 14.80
& 39.02 & 29.88 & 15.24
& 35.00 & 27.40 & 18.00 \\
\rowcolor{gray!15}
\cellcolor{white}
& Ours
& 71.80 & \textbf{67.63} & \textbf{47.99}
& \textbf{24.00} & \textbf{20.40} & \textbf{15.00}
& \textbf{43.90} & \textbf{33.54} & \textbf{18.90}
& \textbf{35.80} & \textbf{29.00} & \textbf{18.40} \\

\midrule

\multirow{3}{*}{LLaDA}
& Vanilla 
& 71.65 & 64.44 & \textbf{36.77}
& \textbf{24.40} & 18.20 & 12.80
& 39.02 & 24.39 & 15.24
& 33.20 & 27.00 & 16.00 \\
& Euclidean
& 72.33 & 64.29 & 36.69
& 22.60 & 19.40 & \textbf{14.80}
& 41.47 & 31.71 & 15.24
& 34.60 & 26.80 & 17.80 \\
\rowcolor{gray!15}
\cellcolor{white}
& Ours
& \textbf{73.24} & \textbf{65.58} & \textbf{36.77}
& 23.60 & \textbf{20.00} & \textbf{14.80}
& \textbf{45.12} & \textbf{32.32} & \textbf{19.51}
& \textbf{36.20} & \textbf{28.60} & \textbf{19.20} \\

\midrule

\multirow{3}{*}{Dream}
& Vanilla 
& 74.45 & 56.94 & 16.15
& 20.40 & 5.20 & 0.40
& 42.68 & 17.68 & 3.05
& 43.40 & 25.20 & 14.20 \\
& Euclidean
& 77.03 & 56.25 & 16.22
& 21.80 & 6.00 & \textbf{0.80}
& 46.34 & 23.78 & 8.54
& 45.00 & 24.80 & 15.00 \\
\rowcolor{gray!15}
\cellcolor{white}
& Ours
& \textbf{78.17} & \textbf{58.61} & \textbf{18.20}
& \textbf{22.40} & \textbf{6.40} & \textbf{0.80}
& \textbf{50.61} & \textbf{32.32} & \textbf{14.02}
& \textbf{47.00} & \textbf{32.80} & \textbf{17.40} \\
\bottomrule
\end{tabular}
}
\vspace{-10pt}
\end{table*}

\subsection{Main Results}
\label{sec:main_results}
Table~\ref{tab:main_accuracy} compares our method with standard parallel decoding without soft-token feedback (Vanilla) and the training-free soft-token baseline using linear interpolation in embedding space (Euclidean).
Our method outperforms both baselines in most settings without additional training.

On GSM8K with LLaDA-2.0 mini under the same decoding setting, our method achieves 87.40\% accuracy, yielding gains of 4.31 and 1.58 percentage points over the respective baselines.
Gains extend to other decoding settings.
On HumanEval with Dream at four tokens per step, our method achieves 32.32\% accuracy, outperforming vanilla parallel decoding and Euclidean soft tokens by 14.64 and 8.54 percentage points, respectively.
These results demonstrate that our soft-token construction can improve parallel decoding in frozen pretrained DLMs.
Appendix~\ref{app:additional_results} presents an ablation study, hyperparameter sensitivity analyses, and results showing that our method can improve existing adaptive decoding methods~\citep{wu2025fast,benhamu2025accelerated}.

\begin{table}[t]
\centering
\small
\caption{
Embedding geometry on GSM8K and HumanEval.
Candidate--candidate reports the mean pairwise cosine similarity among the top-three predicted-token embeddings.
Aggregate--\texttt{[MASK]} reports the mean absolute cosine similarity between the probability-weighted average of these embeddings and the \texttt{[MASK]} embedding.
}
\vspace{-8pt}
\label{tab:top3_geometry}

\begin{tabular}{llcc}
\toprule
Model & Dataset & Candidate--candidate & Aggregate--\texttt{[MASK]} \\
\midrule
\multirow{2}{*}{LLaDA}
    & GSM8K     & 0.496 & 0.015 \\
    & HumanEval & 0.447 & 0.011 \\
\midrule
\multirow{2}{*}{LLaDA-1.5}
    & GSM8K     & 0.543 & 0.014 \\
    & HumanEval & 0.497 & 0.011 \\
\midrule
\multirow{2}{*}{LLaDA-2.0 mini}
    & GSM8K     & 0.724 & 0.043 \\
    & HumanEval & 0.629 & 0.074 \\
\midrule
\multirow{2}{*}{Dream}
    & GSM8K     & 0.566 & 0.028 \\
    & HumanEval & 0.750 & 0.070 \\
\bottomrule
\end{tabular}
\vspace{-8pt}
\end{table}

\subsection{Geometric Analysis of Token Embeddings}
\label{sec:geometry_analysis}
To examine the geometry motivating our construction in Section~\ref{subsec:mismatch}, we analyze four DLMs on GSM8K and HumanEval.
For each evaluated unresolved token position in the output sequence, we consider the top-three candidate embeddings and measure two quantities: (i) their average pairwise cosine similarity and (ii) the absolute cosine similarity between their probability-weighted aggregate and the \texttt{[MASK]} embedding.

Table~\ref{tab:top3_geometry} shows that candidate embeddings are more aligned with one another, on average, than their aggregate is with the \texttt{[MASK]} embedding across all eight model--dataset pairs. 
This finding supports retaining Euclidean averaging as a practical choice for candidate aggregation.

The low absolute cosine similarities between the aggregate and \texttt{[MASK]} indicate a large angular separation.
Linear interpolation between such endpoints can distort angular information and reduce the resulting embedding norm.
These properties motivate our use of spherical interpolation to control the transition in direction while separately preserving the \texttt{[MASK]} embedding norm.

\begin{table}[t]
\centering
\caption{
Reverse KL divergence from the factorized joint prediction to the dependency-preserving reference.
The lowest reverse KL for each model is shown in bold.
}
\vspace{-8pt}
\label{tab:sequence_reverse_kl}
\setlength{\tabcolsep}{10pt}
\begin{tabular}{lccc}
\toprule
Model & Vanilla  & Euclidean & Ours \\
\midrule
LLaDA          & 2.1989 & 0.3415 & \textbf{0.2458} \\
LLaDA-1.5      & 2.0327 & 1.9273 & \textbf{0.5624} \\
LLaDA-2.0 mini & 2.7338 & 1.5610 & \textbf{1.1739} \\
Dream          & 4.2733 & 4.1139 & \textbf{1.7553} \\
\bottomrule
\end{tabular}
\vspace{-10pt}
\end{table}

\subsection{Analyzing Agreement-Seeking Behavior}
\label{sec:mechanism_analysis}
\vspace{-2pt}
\paragraph{Comparison with additive and multiplicative mixtures.}
Constructing the additive and multiplicative reference mixtures in Sections~\ref{sec:preserve_uncertainty} and~\ref{sec:mixture_behavior} requires evaluating the model under every joint discrete realization of the soft-token inputs.
The number of realizations grows exponentially with the number of soft-token positions, making exhaustive evaluation impractical for long sequences.
We therefore use 70 controlled prompts containing two to four masked positions, where individually plausible predictions can form inconsistent combinations.
For example, a prompt that admits either \emph{New York} or \emph{San Diego} can yield incompatible combinations such as \emph{New Diego} when the two positions are predicted independently.
We measure the Jensen--Shannon (JS) divergence between each reference mixture and the predictive distributions induced by our construction, with results presented in Section~\ref{sec:mixture_behavior}.
Details of the prompts and implementation are provided in Appendix~\ref{app:toy}.

\paragraph{Sequence-level consistency.}
We assess sequence-level consistency using the same toy examples as in the preceding mixture analysis.
We obtain per-position predictive distributions from masked inputs for vanilla parallel decoding and from the corresponding soft-token inputs for the two soft-token methods.
We compute the joint probability of each complete token combination as the product of its predicted token probabilities across answer positions.
As described in Section~\ref{sec:additive_multiplicative}, we measure reverse KL from this factorized joint to a sequential reference.

Table~\ref{tab:sequence_reverse_kl} shows that our method achieves the lowest reverse KL across all four models.
The Euclidean baseline also yields lower divergence than vanilla parallel decoding, but our method consistently provides a larger reduction.
These results provide complementary evidence for the proposed agreement-seeking behavior, supporting the interpretation that our soft-token feedback favors coherent alternatives over inconsistent combinations in these controlled settings.

\section{Related Work}
\label{sec:related_work}

Diffusion language models (DLMs) have emerged as a promising alternative to autoregressive generation.
D3PM introduced a framework for diffusion over discrete data using structured corruption processes~\citep{austin2021structured}.
Subsequent work advanced discrete diffusion through continuous-time formulations~\citep{campbell2022continuous,sun2022score}, improved learning objectives~\citep{meng2022concrete,benton2022denoising,lou2024discrete}, and accelerated sampling algorithms~\citep{chen2023fast}.
Recent studies simplify and unify masked diffusion formulations, improving efficiency and scalability for language modeling~\citep{sahoo2024simple,shi2024simplified}.
At larger scales, LLaDA demonstrates competitive language modeling and instruction following through training from scratch~\citep{nie2025large}, while Dream adapts pretrained autoregressive models to diffusion-based generation~\citep{ye2025dream}.
LLaDA 1.5 improves alignment through variance-reduced preference optimization~\citep{zhu2026llada}, and LLaDA 2.0 scales diffusion language models to 100B total parameters through progressive conversion of pretrained autoregressive models~\citep{bie2025llada}.

Soft tokens and continuous feedback have been explored in autoregressive models to support reasoning beyond discrete token sequences.
Existing approaches use hidden-state feedback~\citep{hao2025training,shen2025codi}, learned soft thought representations~\citep{xu2025softcot,xu2025softcotpp}, compressed latent reasoning states~\citep{tan2025think}, or probability-weighted embedding mixtures~\citep{zhang2025soft}.
Inspired by the success of continuous feedback in autoregressive models, recent work incorporates continuous feedback into DLMs.
Soft-Masking blends the \texttt{[MASK]} embedding with predicted-token embeddings and trains models to process these inputs~\citep{hersche2026soft}.
EvoToken-DLM introduces progressive soft-token refinement supported by continuous trajectory supervision~\citep{zhong2026beyond}, while DMax combines on-policy uniform training with soft parallel decoding to enable iterative revision in embedding space~\citep{chen2026dmax}.
We study soft-token feedback in frozen pretrained DLMs, developing a geometry-aware construction and examining how feedback reshapes predictive distributions and improves sequence-level consistency without additional training.

Concurrent work~\citep{nigam2026lost} uses spherical interpolation with iterative Riemannian candidate aggregation, evaluated through continued pretraining of a 169M-parameter model. 
We develop a closed-form, training-free construction supported by geometric analysis across four pretrained DLMs and investigate its effects on predictive distributions and sequence-level consistency.

\section{Limitations and Conclusion}

\paragraph{Limitation.}
Our exact comparison with additive and multiplicative reference mixtures in Section~\ref{sec:mechanism_analysis} is limited to controlled prompts with two to four masked positions rather than sequences from real-world data.
However, constructing the additive and multiplicative reference mixtures requires enumerating all joint discrete realizations, whose number grows exponentially with the number of soft-token positions.
This makes exhaustive evaluation computationally infeasible for the longer sequences encountered in real-world datasets.
The controlled setting enables exact evaluation of these mixtures while making dependencies and inconsistent token combinations explicit. 

\paragraph{Conclusion.}
We introduced an agreement-seeking view of soft-token feedback for parallel decoding.
Our probabilistic analysis connects multiplicative-mixture behavior with improved sequence-level consistency, supporting the interpretation that feedback favors compatible predictions before commitment.
To study this behavior without additional training, we developed a geometry-aware soft-token method that outperforms standard parallel decoding and a training-free Euclidean baseline across four frozen pretrained DLMs and four math and code benchmarks.
These findings extend the understanding of soft tokens beyond uncertainty preservation.
We believe this probabilistic perspective can help design more effective representations for parallel decoding.

\section*{Ethics Statement}

This work studies decoding methods for pretrained diffusion language models using public benchmarks and synthetic prompts. 
Our method does not address biases, harmful outputs, or potential misuse of the underlying models. 
These concerns remain relevant when deploying models with the proposed decoding method. 

\section*{Reproducibility Statement}

We describe the evaluated models, benchmarks, and baselines in Section~\ref{sec:setup}, with additional evaluation and decoding details in Appendix~\ref{app:settings}.
Appendix~\ref{app:toy} documents the controlled experiments and provides the complete toy prompt collection.

\section*{AI Use Statement.}
We used generative AI tools to improve the clarity and readability of the manuscript and to generate toy datasets for analyzing our method in Section~\ref{sec:mechanism_analysis}. We reviewed the generated examples and AI-assisted revisions and take responsibility for the final content of this work.

\bibliographystyle{iclr2027_conference}
\bibliography{iclr2027_conference}

\newpage

\appendix

\section*{Appendix}

This appendix is organized as follows.
Appendix~\ref{app:settings} provides additional evaluation and decoding details.
Appendix~\ref{app:additional_results} presents additional analyses and ablation study.
Appendix~\ref{app:recovery} reports recovery rates relative to single-token prediction.
Appendix~\ref{app:toy} describes the controlled toy experiments and provides the complete prompt collection.

\section{Experimental Details}
\label{app:settings}

This section supplements the experimental settings in Section~\ref{sec:setup}.

\subsection{Evaluation Setup}
\label{app:evaluation_setup}

We conduct all experiments using the dLLM framework~\citep{zhou2026dllm} and follow the task configurations implemented in lm-evaluation-harness~\citep{gao2024language}.

For LLaDA-Instruct-8B, LLaDA-1.5-Instruct, and LLaDA-2.0-mini, we use 5-shot evaluation on GSM8K, 4-shot on MATH500, 0-shot on HumanEval, and 3-shot on MBPP.
We generate up to 512 tokens on GSM8K, MATH500, and HumanEval, and 256 tokens on MBPP.

For Dream-7B, we follow its default zero-shot configuration on all four benchmarks, with maximum generation lengths of 256, 512, 768, and 1,024 tokens for GSM8K, MATH500, HumanEval, and MBPP, respectively.
We apply each model's chat template and do not use block-diffusion sampling.

\subsection{Decoding Configurations}
\label{app:decoding_configurations}
\paragraph{Decoding protocol.}
All model parameters remain frozen.
We compare vanilla parallel decoding, the Euclidean soft-token baseline, and our geometry-aware construction at 2, 4, and 8 tokens committed per denoising iteration.
Vanilla decoding retains discrete \texttt{[MASK]} tokens at unresolved positions, whereas the soft-token methods feed updated continuous representations into the next iteration.
We report exact-match accuracy for GSM8K and MATH500 and pass@$1$ for HumanEval and MBPP. 

\paragraph{Hyperparameter settings.}
Both soft-token methods use the same candidate aggregation, decoding schedule, and hyperparameter search space.
We search over $\lambda \in \{0.1, 0.3, 0.5, 0.7\}$ and $k \in \{2, 3, 4\}$.
We configure our method for mathematical reasoning and code generation, keeping the settings fixed across benchmarks and parallel-decoding rates within each domain.
For the LLaDA family, we use $\lambda=0.3$ in both domains, with $k=2$ for mathematical reasoning and $k=3$ for code generation.
For Dream, we use $(k,\lambda)=(3,0.1)$ for mathematical reasoning and $(k,\lambda)=(2,0.3)$ for code generation.
We use spherical interpolation and apply soft-token feedback throughout the denoising process.
Appendix~\ref{app:hyperparameters} analyzes sensitivity to $k$ and $\lambda$.

\section{Additional Analyses and Ablations}
\label{app:additional_results}

\subsection{Geometric Limitations of Linear Interpolation from \texttt{[MASK]}}
\label{sec:mask-interpolation-geometry}

To examine the geometric effects discussed in
Section~\ref{subsec:mismatch}, we evaluate LLaDA-8B-Instruct on GSM8K and HumanEval.
At each masked position, we construct a probability-weighted aggregate $\mathbf{m}$ of the top-four predicted token embeddings,
with probabilities renormalized over these candidates, and compute
\[
\mathbf{e}_{\mathrm{linear}}
=
(1-\lambda)\mathbf{e}_{\mathrm{MASK}}
+
\lambda\mathbf{m}.
\]
We use $\lambda=0.5$, for which spherical interpolation between the normalized endpoint directions reaches the angular midpoint.

The candidate aggregate and the \texttt{[MASK]} embedding were nearly orthogonal, with a mean angular separation of $89.42^\circ$.
Linear interpolation traversed, on average, $59.9\%$ of this separation rather than $50\%$.
Its output norm averaged $0.863$ times the \texttt{[MASK]} norm and fell below both endpoint norms at $99.4\%$ of positions.

These measurements show that equal Euclidean interpolation weights do not generally yield the angular midpoint and that interpolation can shrink the representation below both endpoint norms.
They support controlling direction and magnitude separately when incorporating candidate information into the \texttt{[MASK]} representation.

\subsection{Effects of Interpolation and Norm Preservation}
\label{app:mask_norm}

Our construction uses spherical interpolation (SLERP) to determine the
soft token's direction and the \texttt{[MASK]} embedding norm to set its
magnitude.
We evaluate two variants to examine these choices separately.
First, we retain SLERP but replace the output norm
$\lVert\mathbf{e}_{\mathrm{MASK}}\rVert_2$ with
$\lVert\mathbf{m}\rVert_2$, where $\mathbf{m}$ is the aggregated
candidate representation defined in Section~\ref{subsec:construction}.
Second, we replace SLERP with linear interpolation while preserving
the \texttt{[MASK]} embedding norm:
\[
\mathbf{e}_{\mathrm{linear}}
=
\lVert\mathbf{e}_{\mathrm{MASK}}\rVert_2
\frac{
(1-\lambda)\mathbf{e}_{\mathrm{MASK}}+\lambda\mathbf{m}
}{
\lVert(1-\lambda)\mathbf{e}_{\mathrm{MASK}}+\lambda\mathbf{m}\rVert_2
}.
\]
The first variant changes only the output magnitude, while the second
changes only the interpolation rule used to determine the direction.

As shown in Table~\ref{tab:mask_norm_ablation}, our construction achieves
the highest accuracy across all six model--benchmark pairs.
The decrease under candidate-norm scaling supports preserving the
\texttt{[MASK]} embedding norm, while the gap to norm-matched linear
interpolation suggests that norm preservation alone does not account
for the observed gains.

\begin{table}[!htbp]
\centering
\caption{
Effects of interpolation and output norm on accuracy (\%).
SLERP with the \texttt{[MASK]} norm is our standard construction.
The best result for each model--benchmark pair is shown in bold.
}
\label{tab:mask_norm_ablation}
\small
\setlength{\tabcolsep}{4pt}
\renewcommand{\arraystretch}{1.1}
\begin{tabular}{ccccc}
\toprule
& & \multicolumn{2}{c}{SLERP} & Linear \\
\cmidrule(lr){3-4}
\cmidrule(lr){5-5}
Model & Benchmark
& \texttt{[MASK]} norm
& Candidate norm
& \texttt{[MASK]} norm \\
\midrule
Dream
& MATH-500 & \textbf{22.40} & 21.20 & 16.40 \\
& HumanEval & \textbf{50.61} & 50.00 & 50.00 \\
\midrule
LLaDA
& MATH-500 & \textbf{23.60} & 22.60 & 21.80 \\
& HumanEval & \textbf{45.12} & 41.46 & 42.07 \\
\midrule
LLaDA 1.5
& MATH-500 & \textbf{24.00} & 21.80 & 21.20 \\
& HumanEval & \textbf{43.90} & 39.63 & 37.80 \\
\bottomrule
\end{tabular}
\end{table}

\subsection{Interpolation Strength and Candidate Count}
\label{app:hyperparameters}

\begin{figure}[!htbp]
    \centering
    \begin{subfigure}[t]{0.49\textwidth}
        \centering
        \includegraphics[width=\linewidth]{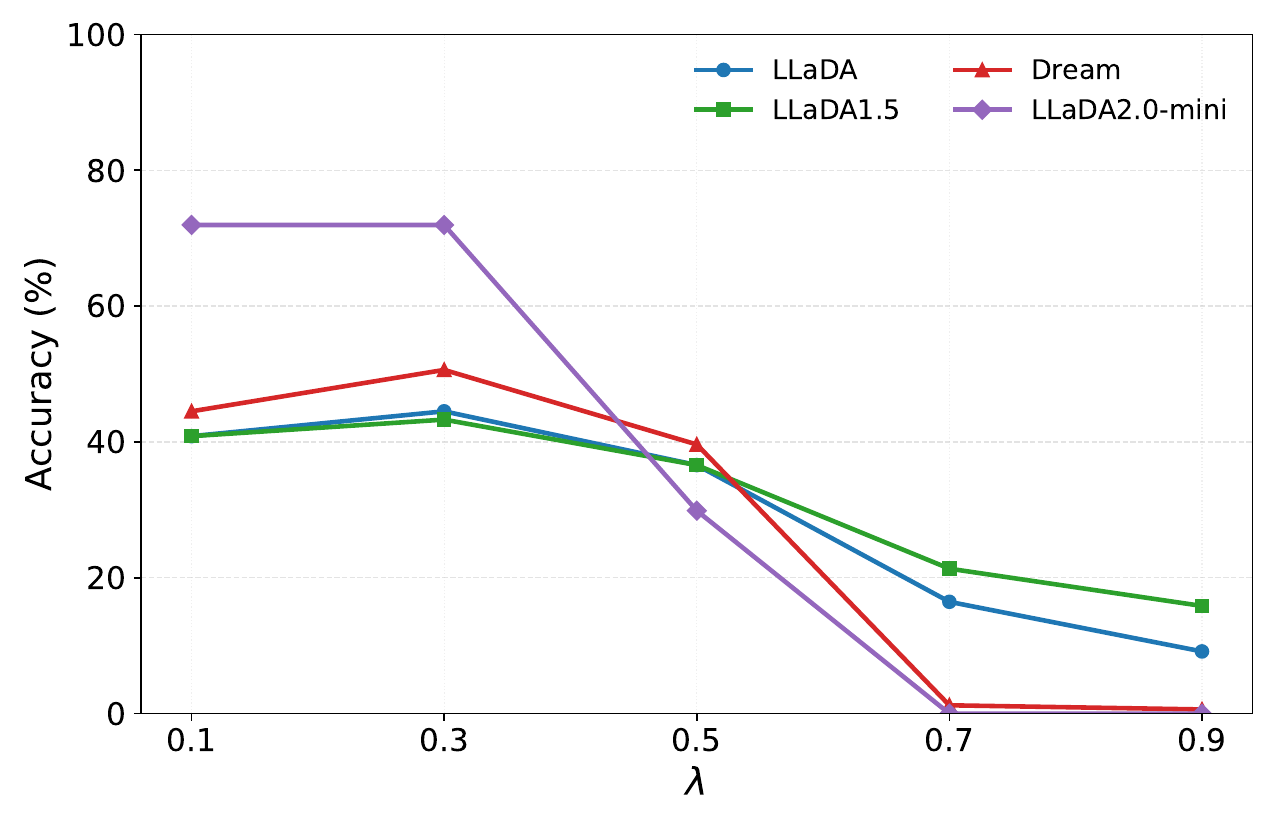}
        \caption{Interpolation coefficient $\lambda$.}
        \label{fig:hparam-lambda}
    \end{subfigure}
    \hfill
    \begin{subfigure}[t]{0.49\textwidth}
        \centering
        \includegraphics[width=\linewidth]{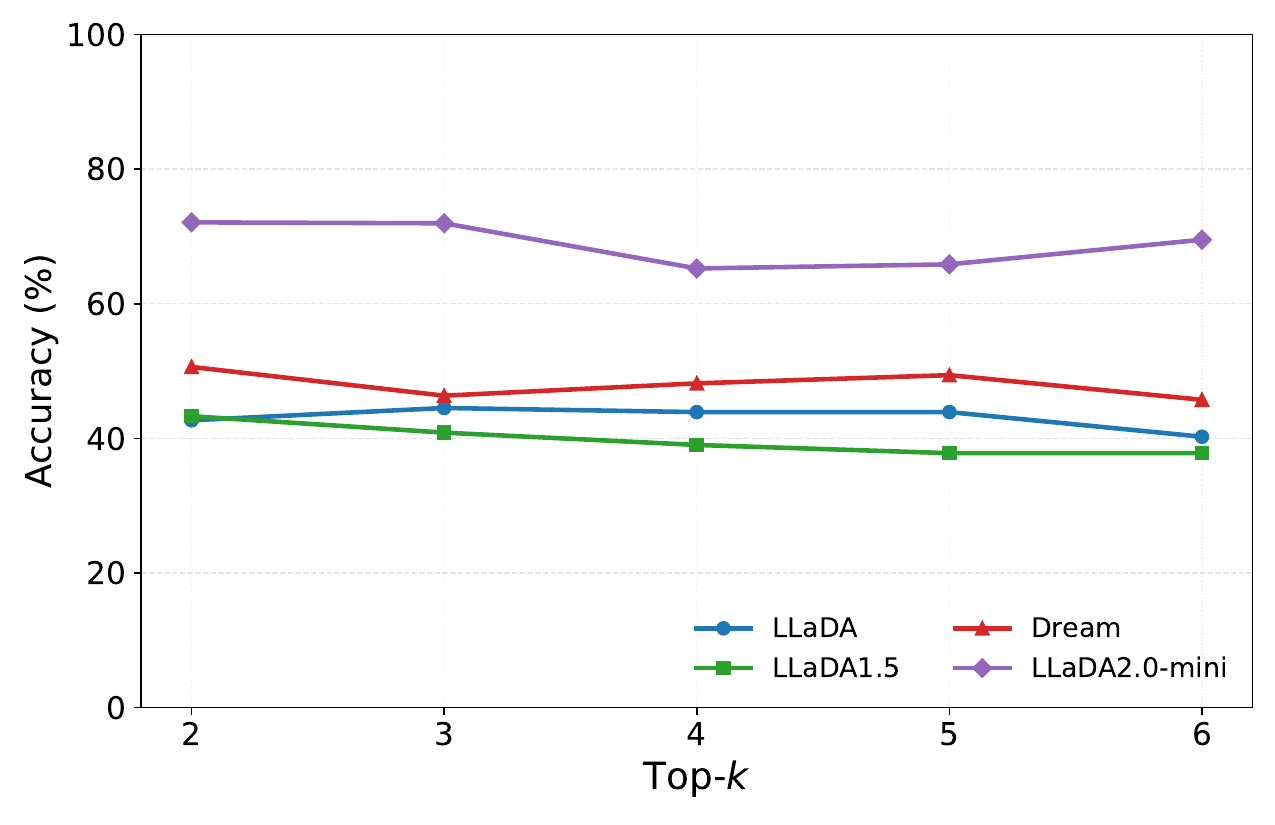}
        \caption{Number of candidate tokens $k$.}
        \label{fig:hparam-top-k}
    \end{subfigure}
    \caption{
        Sensitivity to interpolation strength and candidate count
        in soft-token decoding.
    }
    \label{fig:hparam}
\end{figure}

The interpolation coefficient $\lambda$ controls how far the soft-token direction moves from \texttt{[MASK]} toward the candidate aggregate, while $k$ determines how many candidate embeddings contribute to that aggregate.
Figure~\ref{fig:hparam} examines sensitivity to these two parameters on HumanEval.

\paragraph{Interpolation strength.}
Smaller values of $\lambda$ keep the soft-token embedding closer to \texttt{[MASK]}, whereas larger values give greater influence to the candidate aggregate.
Across the four evaluated models, $\lambda=0.3$ achieves the best results, while larger values reduce accuracy, with the severity of the decline varying across models.
We hypothesize that strong interpolation moves the embedding too far from \texttt{[MASK]}.
Our construction is intended to enrich an unresolved position with predictive information while retaining its role as a masked position.
Excessive deviation from the pretrained \texttt{[MASK]} representation may make it harder for the frozen model to interpret the position as unresolved, degrading subsequent predictions.
These results favor moderate interpolation that incorporates candidate information while maintaining proximity to \texttt{[MASK]}.

\paragraph{Candidate count.}
Accuracy is comparatively stable across the evaluated candidate counts, and increasing $k$ does not consistently improve performance.
A plausible explanation is that probability weighting limits the influence of additional, lower-probability candidates on the aggregated representation.
This limited sensitivity makes the method less dependent on precise tuning of the candidate count.

\subsection{Combining Soft Tokens with Adaptive Sampling}
\label{sec:adaptive_sampling}

We further investigate whether our soft-token feedback complements adaptive sampling strategies.
Specifically, we incorporate our method into confidence-threshold decoding from Fast-dLLM~\citep{wu2025fast} and entropy-bounded unmasking in EB-Sampler~\citep{benhamu2025accelerated}.
We set the confidence threshold to $0.9$ and the entropy bound to $0.1$.
These strategies adaptively determine which tokens to commit at each step, while our method refines the predictive distributions through soft-token feedback.
We evaluate each sampler with and without our feedback on GSM8K and HumanEval using LLaDA and Dream.

As shown in Table~\ref{tab:adaptive_sampling}, our method improves both samplers across all evaluated model--dataset combinations, with gains ranging from 0.40 to 3.66 percentage points.
These results suggest that our soft-token feedback complements adaptive token selection and that its benefits extend beyond fixed-size parallel decoding.

\begin{table}[t]
\centering
\small
\caption{
Combining our soft-token feedback with adaptive samplers.
We report GSM8K accuracy and HumanEval pass@1 (\%; higher is better).
Confidence thresholding uses the decoding rule from Fast-dLLM.
Bold values indicate improvements over the corresponding sampler without our feedback.
}
\label{tab:adaptive_sampling}
\setlength{\tabcolsep}{7pt}
\begin{tabular}{@{}lcc@{\hspace{1.5em}}cc@{}}
\toprule
& \multicolumn{2}{c}{LLaDA}
& \multicolumn{2}{c}{Dream} \\
\cmidrule(lr){2-3}\cmidrule(lr){4-5}
Sampler & GSM8K & HumanEval & GSM8K & HumanEval \\
\midrule
Confidence thresholding
& 29.80 & 45.12 & 79.80 & 55.49 \\
\quad + Ours
& \textbf{30.20} & \textbf{48.78}
& \textbf{81.60} & \textbf{58.54} \\
\addlinespace
EB-Sampler
& 24.00 & 46.34 & 82.60 & 56.71 \\
\quad + Ours
& \textbf{26.00} & \textbf{48.17}
& \textbf{83.60} & \textbf{58.54} \\
\bottomrule
\end{tabular}
\end{table}

\section{Recovery of Single-Token Predictions}
\label{app:recovery}

Beyond aggregate accuracy, we examine whether parallel decoding preserves correctness on examples solved by single-token prediction (STP), which commits one token per denoising iteration.
Similar aggregate accuracy does not necessarily imply such preservation.
A parallel method can match STP's accuracy while failing on many examples that STP solves, with these losses offset by successes on other examples.
We observe this discrepancy in some settings, motivating a direct assessment of how well parallel decoding retains STP's successes.

Let $S_{\mathrm{STP}}$ and $S_{\mathrm{parallel}}$ denote the sets of examples correctly solved by STP and a given parallel decoding method, respectively.
We define recovery rate as
\[
\operatorname{Recovery\ rate}
=
100\times
\frac{
    |S_{\mathrm{STP}}\cap S_{\mathrm{parallel}}|
}{
    |S_{\mathrm{STP}}|
}.
\]
This metric measures the percentage of examples solved by STP that remain correctly solved under parallel decoding.
It gives no credit for examples solved only by the parallel method, and therefore complements overall accuracy.

Table~\ref{tab:main_recovery} reports recovery rates for mathematical reasoning and code generation.
Recovery rates decrease as more tokens are committed per iteration, indicating that greater parallelism makes it harder to retain correctness on examples solved by STP.
Our method achieves the highest recovery rate in most settings, suggesting that its accuracy gains are accompanied by better retention of STP-correct examples.
For example, on HumanEval at eight tokens per step, our method improves recovery over vanilla parallel decoding from 22.78\% to 37.97\% for LLaDA-1.5 and from 5.26\% to 24.21\% for Dream.
Nevertheless, recovery remains low in some highly parallel settings, indicating that soft-token feedback mitigates but does not eliminate the degradation associated with parallel decoding.

\begin{table}[!htbp]
\centering
\small
\caption{
Recovery rate (\%) on mathematical reasoning and code generation.
Columns labeled 2, 4, and 8 indicate tokens committed per denoising iteration.
The best result for each model, benchmark, and decoding setting is in bold.
}
\vspace{-8pt}
\label{tab:main_recovery}
\setlength{\tabcolsep}{4pt}
\resizebox{\textwidth}{!}{%
\begin{tabular}{ll*{12}{r}}
\toprule
& & \multicolumn{3}{c}{GSM8K}
& \multicolumn{3}{c}{MATH500}
& \multicolumn{3}{c}{HumanEval}
& \multicolumn{3}{c}{MBPP} \\
\cmidrule(lr){3-5}
\cmidrule(lr){6-8}
\cmidrule(lr){9-11}
\cmidrule(lr){12-14}
Model & Method
& 2 & 4 & 8
& 2 & 4 & 8
& 2 & 4 & 8
& 2 & 4 & 8 \\
\midrule

\multirow{3}{*}{LLaDA-2.0 mini}
& Vanilla
& 95.50 & 88.33 & 53.67
& 88.83 & 82.04 & 37.38
& 83.33 & 56.82 & \textbf{28.79}
& 83.70 & 64.89 & 34.99 \\
& Euclidean
& 95.41 & 91.92 & 55.24
& 92.72 & 81.07 & 41.26
& 81.06 & 62.12 & 28.03
& \textbf{85.89} & \textbf{66.46} & 34.80 \\
\rowcolor{gray!15}
\cellcolor{white}
& Ours
& \textbf{96.94} & \textbf{93.23} & \textbf{58.73}
& \textbf{94.17} & \textbf{85.92} & \textbf{42.23}
& \textbf{85.61} & \textbf{64.39} & 28.03
& 85.58 & \textbf{66.46} & \textbf{35.42} \\

\midrule

\multirow{3}{*}{LLaDA-1.5}
& Vanilla
& \textbf{91.25} & 83.35 & 56.59
& 71.31 & \textbf{59.84} & 40.00
& 70.89 & 46.84 & 22.78
& 78.50 & 61.00 & 38.50 \\
& Euclidean
& 89.99 & 84.62 & 61.01
& 68.85 & 59.02 & 46.72
& 69.62 & 54.43 & 26.58
& 78.50 & 59.50 & \textbf{40.50} \\
\rowcolor{gray!15}
\cellcolor{white}
& Ours
& 90.83 & \textbf{85.67} & \textbf{61.22}
& \textbf{76.23} & \textbf{59.84} & \textbf{47.54}
& \textbf{74.68} & \textbf{59.49} & \textbf{37.97}
& \textbf{79.00} & \textbf{62.00} & 40.00 \\

\midrule

\multirow{3}{*}{LLaDA}
& Vanilla
& 89.24 & 78.07 & \textbf{45.89}
& 67.19 & 51.56 & 38.28
& 76.32 & 48.68 & 30.26
& 73.00 & 61.00 & 38.00 \\
& Euclidean
& 89.34 & 79.19 & 45.79
& 65.63 & \textbf{55.47} & 40.63
& 75.00 & 59.21 & 28.95
& 75.50 & 58.00 & 38.00 \\
\rowcolor{gray!15}
\cellcolor{white}
& Ours
& \textbf{90.05} & \textbf{80.41} & \textbf{45.89}
& \textbf{71.88} & \textbf{55.47} & \textbf{41.41}
& \textbf{85.53} & \textbf{60.53} & \textbf{35.53}
& \textbf{78.50} & \textbf{64.00} & \textbf{44.50} \\

\midrule

\multirow{3}{*}{Dream}
& Vanilla
& 84.74 & 65.03 & 18.41
& 42.37 & 11.70 & 0.58
& 66.32 & 29.47 & 5.26
& 70.45 & 39.86 & 23.71 \\
& Euclidean
& 86.68 & 64.20 & 18.32
& 42.69 & \textbf{13.45} & 0.59
& 72.63 & 38.95 & 13.68
& 70.80 & 40.50 & 24.41 \\
\rowcolor{gray!15}
\cellcolor{white}
& Ours
& \textbf{87.60} & \textbf{66.98} & \textbf{20.44}
& \textbf{46.20} & \textbf{13.45} & \textbf{1.17}
& \textbf{74.74} & \textbf{54.74} & \textbf{24.21}
& \textbf{74.23} & \textbf{53.61} & \textbf{28.09} \\

\bottomrule
\end{tabular}%
}
\end{table}

\section{Supplementary Analysis of Soft-Token Feedback}
\label{app:toy}

This appendix supplements the analysis in Section~\ref{sec:mechanism_analysis}.
We use controlled prompts to compare soft-token predictions with additive and multiplicative reference mixtures and to evaluate agreement with a sequential reference.
We then extend the reverse-KL comparison to model-generated completions.
The complete controlled prompt collection is provided in Appendix~\ref{app:toy_catalogue}.

\subsection{Prompt Design}
\label{app:toy_design}

We use 70 controlled prompts: 30 with two mask placeholders, 20 with three, and 20 with four.
Each prompt specifies two alternative answers.
For example, a two-mask prompt asks for either \emph{New York} or \emph{San Diego}.
Independent predictions can mix these alternatives to produce \emph{New Diego} or \emph{San York}, violating the stated choice.
These prompts therefore expose dependencies between predicted positions and allow us to examine whether soft-token feedback favors mutually consistent predictions.

We use the same prompt collection for reference-mixture comparison and the sequence-level consistency analysis.
The prompts were generated with assistance from generative AI and reviewed by the authors, as described in the AI Use Statement.
The complete collection appears in Tables~\ref{tab:toy_2}--\ref{tab:toy_4}.

\subsection{Comparison with Reference Mixtures}
\label{app:toy_mixtures}

We compare the predictive distributions induced by Euclidean soft tokens and our construction with the additive and multiplicative references defined in Section~\ref{sec:preserve_uncertainty}.
We use the controlled prompts described in Appendix~\ref{app:toy_design}.

\paragraph{Soft-token predictions.}
For each prompt, we first evaluate the frozen model with all answer positions masked.
We use the resulting predictive distributions to select candidate tokens and construct the soft-token embeddings.
The continuous-input predictions are obtained from one additional forward pass with these embeddings, without committing discrete tokens or performing further feedback updates.
We use $k=3$ and $\lambda=0.3$ for both soft-token methods.
The initial and feedback predictive distributions are computed using the model's unscaled softmax.

\paragraph{Discrete reference construction.}
For a queried position $j$, let $\mathcal{S}_j$ denote the set of input positions represented by soft tokens in the comparison.
At each position in $\mathcal{S}_j$, we consider the selected candidate tokens and the unresolved \texttt{[MASK]} state.
Let $\mathcal{Z}_j$ denote the Cartesian product of these positionwise state sets.
With $k$ candidates per position, this gives up to
$(k+1)^{|\mathcal{S}_j|}$ joint realizations.
We enumerate all realizations, keeping the prompt and the remaining input positions fixed.

For each queried answer position $j$, $\mathcal{S}_j$ contains
the positions represented by soft tokens in the comparison,
including $j$ itself.
We enumerate their joint discrete realizations by replacing
each soft-token embedding with either the \texttt{[MASK]}
embedding or one of its selected candidate-token embeddings.
The prompt and all inputs outside $\mathcal{S}_j$ remain fixed.

Let $w_j(\mathbf{z})$ be the weight of realization
$\mathbf{z}\in\mathcal{Z}_j$, obtained by multiplying the positionwise realization weights defined in Section~\ref{sec:preserve_uncertainty}.
Let $P_{j,\mathbf{z}}$ denote the model's predictive distribution at position $j$ under this discrete input.
The two references are
\[
P_{\mathrm{add},j}(y)
=
\sum_{\mathbf{z}\in\mathcal{Z}_j}
w_j(\mathbf{z})P_{j,\mathbf{z}}(y),
\]
and
\[
P_{\mathrm{mult},j}(y)
=
\frac{
\exp\!\left(
\sum_{\mathbf{z}\in\mathcal{Z}_j}
w_j(\mathbf{z})\log P_{j,\mathbf{z}}(y)
\right)
}{
\displaystyle
\sum_{y'}
\exp\!\left(
\sum_{\mathbf{z}\in\mathcal{Z}_j}
w_j(\mathbf{z})\log P_{j,\mathbf{z}}(y')
\right)
}.
\]

The candidate probabilities are taken from the initial
full-vocabulary distribution and renormalized over the selected
top-$k$ candidates.
Writing these normalized probabilities as $\bar p_i$, the
positionwise realization weights are $1-\lambda$ for
\texttt{[MASK]} and $\lambda\bar p_i$ for candidate $i$.
The additive and multiplicative references use identical candidate
sets, discrete realizations, and realization weights.
They differ only in their pooling operation: a weighted arithmetic
mean for the additive reference and a normalized weighted geometric
mean for the multiplicative reference.
Discrete-input predictive distributions are computed using the
model's softmax without temperature scaling.

\paragraph{Divergence and aggregation.}
For each method and queried position, we compare the continuous-input prediction with each reference using Jensen--Shannon divergence:
\[
\operatorname{JS}(P,Q)
=
\frac{1}{2}D_{\mathrm{KL}}(P\|M)
+
\frac{1}{2}D_{\mathrm{KL}}(Q\|M),
\qquad
M=\frac{P+Q}{2}.
\]

We compute JS divergence over the full output vocabulary using
natural logarithms.
We first average the divergences over queried positions within
each prompt, then average over prompts, giving each prompt
equal weight regardless of its number of answer positions.

Table~\ref{tab:additive_multiplicative_js} reports the resulting divergences for each model.
Lower values indicate closer agreement with the corresponding reference.

\subsection{Sequence-Level Consistency}
\label{app:toy_consistency}

Following Section~\ref{sec:additive_multiplicative}, we evaluate whether soft-token feedback brings factorized parallel predictions closer to a sequential reference that incorporates dependencies among answer tokens.
Using the controlled prompt collection, we enumerate candidate answer sequences and compare their probabilities under each method and the reference.

\paragraph{Candidate answer support.}
For a prompt with $m$ masked answer positions, let
$\mathbf{a}=(a_1,\ldots,a_m)$ and
$\mathbf{b}=(b_1,\ldots,b_m)$
be the token sequences corresponding to its two stated alternatives. We tokenize each alternative using the evaluated model's tokenizer, preserving the whitespace at the answer boundary. This comparison includes only prompts for which both alternatives occupy exactly $m$ answer tokens.

We define the shared support as
\[
S
=
\left\{
\mathbf{y}\in\mathcal{V}^{m}
:
y_j\in\{a_j,b_j\}
\text{ for every }j
\right\}.
\]
This support contains both stated alternatives and every distinct positionwise recombination of their tokens. Its size is
\[
|S|
=
\prod_{j=1}^{m}|\{a_j,b_j\}|
\leq 2^m.
\]
Thus, two-, three-, and four-token examples contain at most four, eight, and sixteen candidate sequences, respectively. Shared tokens between the alternatives reduce these counts. The support is determined by the stated alternatives and is independent of the top-$k$ candidates used to construct soft-token embeddings.

\paragraph{Order-averaged sequential reference.}
Let $\mathbf{x}$ denote the context with all $m$ answer positions masked, and let $\Pi_m$ be the set of all permutations of these positions. For a candidate sequence $\mathbf{y}\in S$ and an order $\pi=(\pi_1,\ldots,\pi_m)\in\Pi_m$, we score the sequence by teacher forcing. At step $t$, we record the probability of $y_{\pi_t}$ and insert that token before proceeding. Writing $\mathbf{x}_{<t}^{\pi}(\mathbf{y})$ for the context after the first $t-1$ insertions, the joint probability under order $\pi$ is
\[
p_{\pi}(\mathbf{y}\mid\mathbf{x})
=
\prod_{t=1}^{m}
p_{\theta}\!\left(
y_{\pi_t}
\mid
\mathbf{x}_{<t}^{\pi}(\mathbf{y})
\right).
\]
The prompt remains fixed, and answer positions not yet filled remain masked. All conditionals use the model's full-vocabulary softmax without temperature scaling.

We define the sequential reference by uniformly averaging the joint probabilities:
\[
p_{\mathrm{ref}}(\mathbf{y}\mid\mathbf{x})
=
\frac{1}{m!}
\sum_{\pi\in\Pi_m}
p_{\pi}(\mathbf{y}\mid\mathbf{x}).
\]
All two, six, or twenty-four orders are enumerated for $m=2$, $3$, or $4$, respectively. The reference uses discrete token and \texttt{[MASK]} inputs without soft-token feedback and is shared across methods. 

\paragraph{Factorized parallel predictions.}
Vanilla parallel decoding uses the predictive distributions from an initial forward pass with all answer positions masked. 
For Euclidean soft tokens and our construction, these initial predictions are used to construct soft-token embeddings at every answer position. We then obtain updated predictions from one additional forward pass, without committing discrete tokens or performing further feedback updates. Both soft-token methods use $k=3$ and $\lambda=0.3$, with candidate probabilities renormalized over the top-$k$ tokens, as specified in Appendix~\ref{app:toy_mixtures}.

Let $q_j^{(d)}(\cdot\mid\mathbf{x})$ be the resulting predictive distribution at position $j$ for method $d$. Each method assigns a factorized probability to a candidate sequence:
\[
q^{(d)}(\mathbf{y}\mid\mathbf{x})
=
\prod_{j=1}^{m}
q_j^{(d)}(y_j\mid\mathbf{x}).
\]
These probabilities are computed from full-vocabulary softmax distributions without temperature scaling.

\paragraph{Conditional reverse KL.}
We separately normalize each method's prediction and the sequential reference over the shared support:
\[
q_S^{(d)}(\mathbf{y}\mid\mathbf{x})
=
\frac{
q^{(d)}(\mathbf{y}\mid\mathbf{x})
}{
\displaystyle\sum_{\mathbf{z}\in S}
q^{(d)}(\mathbf{z}\mid\mathbf{x})
},
\qquad
p_{\mathrm{ref},S}(\mathbf{y}\mid\mathbf{x})
=
\frac{
p_{\mathrm{ref}}(\mathbf{y}\mid\mathbf{x})
}{
\displaystyle\sum_{\mathbf{z}\in S}
p_{\mathrm{ref}}(\mathbf{z}\mid\mathbf{x})
}.
\]
We then evaluate
\[
D_{\mathrm{KL}}\!\left(
q_S^{(d)}\,\|\,p_{\mathrm{ref},S}
\right)
=
\sum_{\mathbf{y}\in S}
q_S^{(d)}(\mathbf{y}\mid\mathbf{x})
\log
\frac{
q_S^{(d)}(\mathbf{y}\mid\mathbf{x})
}{
p_{\mathrm{ref},S}(\mathbf{y}\mid\mathbf{x})
}.
\]
We use natural logarithms and compute the sum exactly, without sampling candidate sequences or unmasking orders.

Table~\ref{tab:sequence_reverse_kl} reports the mean divergence in nats, giving each evaluated prompt equal weight. Lower values indicate closer agreement with the sequential reference within $S$. Because both distributions are conditioned on this support, the metric does not evaluate their total probability mass on $S$ or discrepancies outside it. The reference is derived from the same model rather than a ground-truth joint distribution, so this comparison serves as a diagnostic of the interpretation in Section~\ref{sec:additive_multiplicative}.

\subsection{Reverse-KL Evaluation on Model-Generated Completions}
\label{app:real_reverse_kl}

To complement the controlled-prompt analysis, we examine agreement with a sequential reference on spans drawn from model-generated completions.
This experiment tests whether the same qualitative pattern extends to mathematical reasoning and code-generation contexts.

\paragraph{Evaluation setup.}
We evaluate spans containing $m\in\{2,3,4\}$ consecutive completion tokens.
Because the number of possible token sequences grows exponentially with span length, we use the finite-support procedure described below.
We use 100 completions for each combination of MATH-500 or HumanEval and Dream-v0-Instruct-7B, LLaDA-1.5, or LLaDA-8B-Instruct.
For each completion, we select two approximately evenly spaced evaluation locations.
The surrounding context, including tokens to the right of each masked span, remains visible.
The evaluation therefore concerns bidirectional infilling with two to four masked tokens.

\paragraph{Compared methods.}
Discrete MTP predicts all masked positions independently from the initial mask-only forward pass.
For the soft-token methods, this pass provides the predictive probabilities, which we renormalize over the top-four candidates to construct
\[
\mathbf{e}_{\mathrm{tok}}
=
\sum_{v\in\operatorname{Top4}(p)}
\frac{p(v)}
{\sum_{u\in\operatorname{Top4}(p)}p(u)}
\mathbf{e}_v.
\]
Linear and SLERP feedback use
\[
\mathbf{e}_{\mathrm{linear}}
=
(1-\lambda)\mathbf{e}_{\mathrm{MASK}}
+
\lambda\mathbf{e}_{\mathrm{tok}},
\]
and
\[
\mathbf{e}_{\mathrm{SLERP}}
=
\lVert\mathbf{e}_{\mathrm{MASK}}\rVert_2
\operatorname{SLERP}\!\left(
\frac{\mathbf{e}_{\mathrm{MASK}}}
     {\lVert\mathbf{e}_{\mathrm{MASK}}\rVert_2},
\frac{\mathbf{e}_{\mathrm{tok}}}
     {\lVert\mathbf{e}_{\mathrm{tok}}\rVert_2};
\lambda
\right),
\]
respectively.
Both soft-token methods use four candidates, $\lambda=0.3$,
and one feedback step.
For each method $a$, the resulting marginals define the
factorized probability of a candidate tuple $z=(z_1,\ldots,z_m)$:
\[
\widetilde p_a(z)
=
\prod_{j=1}^{m}p_a(z_j\mid x),
\]
where $x$ denotes the context with all $m$ span positions masked.

\paragraph{Sequential reference.}
Unlike the order-averaged reference used for the controlled
prompts, this experiment uses confidence-ordered sequential scoring.
For each candidate tuple, we repeatedly select the remaining
masked position with the highest maximum token probability,
score its candidate token, and insert that token before
evaluating the next position.
Let $x_{<t}(z)$ denote the context after inserting the first
$t-1$ selected tokens from $z$, with $x_{<1}(z)=x$, and let
$i_t(z)$ denote the position selected from this context.
The sequential reference is
\[
q_{\mathrm{seq}}(z)
=
\prod_{t=1}^{m}
p_\theta\!\left(
z_{i_t(z)}\mid x_{<t}(z)
\right).
\]
The first selected position is shared across candidate tuples,
whereas subsequent positions may depend on the previously
inserted candidate tokens.

\paragraph{Finite-support evaluation.}
Full-vocabulary enumeration requires evaluating $V^m$ token tuples.
We instead construct a shared candidate set $S_r$ by combining
the top-$r$ Cartesian supports from all three methods with
candidates from confidence-ordered sequential beam search,
using branching factor $r$ and beam width $r^2$.
Special tokens are excluded and duplicate tuples are removed.
The primary evaluation uses $r=4$.

We normalize both distributions on the shared support:
\[
p_{a,S_r}(z)
=
\frac{\widetilde p_a(z)}
     {\sum_{z'\in S_r}\widetilde p_a(z')},
\qquad
q_{S_r}(z)
=
\frac{q_{\mathrm{seq}}(z)}
     {\sum_{z'\in S_r}q_{\mathrm{seq}}(z')}.
\]
We then compute
\[
D_{\mathrm{KL}}\!\left(p_{a,S_r}\,\|\,q_{S_r}\right)
=
\sum_{z\in S_r}
p_{a,S_r}(z)
\log
\frac{p_{a,S_r}(z)}{q_{S_r}(z)}.
\]
This measures reverse KL between distributions conditioned on
$S_r$; discrepancies outside this support are not evaluated.

\paragraph{Aggregation and results.}
We first average the span-level divergences within each
completion and then average over the 100 completions for each
model--dataset pair.

Table~\ref{tab:real_reverse_kl} reports the mean conditional
reverse KL for $r=4$.
SLERP achieves the lowest observed mean in all six settings.
These results are consistent with improved agreement with the
sequential reference within the evaluated support for
bidirectional infilling with two to four masked tokens.
The comparison evaluates the complete feedback constructions,
including their different treatments of interpolation and
output norm.

\begin{table}[t]
\centering
\small
\caption{
Mean conditional reverse KL on the shared support $S_4$, in nats.
Each value averages 100 completion-level measurements.
Lower is better; the lowest mean in each row is bold.
}
\label{tab:real_reverse_kl}
\setlength{\tabcolsep}{7pt}
\begin{tabular}{ccccc}
\toprule
Model & Dataset & SLERP & Linear & Discrete MTP \\
\midrule
Dream
& HumanEval & \textbf{0.000704} & 0.084962 & 0.001848 \\
& MATH-500 & \textbf{0.000554} & 0.003955 & 0.001372 \\
\midrule
LLaDA-1.5
& HumanEval & \textbf{0.035973} & 0.037907 & 0.038491 \\
& MATH-500 & \textbf{0.015964} & 0.016934 & 0.017006 \\
\midrule
LLaDA-8B
& HumanEval & \textbf{0.000665} & 0.000918 & 0.000977 \\
& MATH-500 & \textbf{0.053294} & 0.053845 & 0.065190 \\
\bottomrule
\end{tabular}
\end{table}

\subsection{Complete Prompt Collection}
\label{app:toy_catalogue}

Tables~\ref{tab:toy_2}--\ref{tab:toy_4} list the complete prompt collection, grouped by the number of mask placeholders.
Each table separates a shared prompt template from the variable text, with the caption specifying how to reconstruct each prompt.
The two-mask table includes a lead-in column because its introductory wording varies, whereas the three- and four-mask tables use a fixed lead-in.

\begin{table}[!htbp]
\centering
\small
\caption{
Complete two-mask prompt collection (30 prompts).
Each row instantiates the template:
\emph{\{Lead-in\} either \{Alternative A\} or \{Alternative B\}. The answer is}
\texttt{[MASK][MASK]}.
}
\vspace{-8pt}
\label{tab:toy_2}
\setlength{\tabcolsep}{4pt}
\renewcommand{\arraystretch}{1.08}
\begin{tabular}{
    @{}
    >{\raggedright\arraybackslash}p{0.29\textwidth}
    >{\raggedright\arraybackslash}p{0.25\textwidth}
    >{\raggedright\arraybackslash}p{0.25\textwidth}
    @{}
}
\toprule
Lead-in & Alternative A & Alternative B \\
\midrule
The location is & New York & San Diego \\
The movie series is & Star Wars & Harry Potter \\
The framework is & React Native & Swift UI \\
The dessert is & ice cream & apple pie \\
The pet is & black cat & white dog \\
The meal is & fried rice & tomato soup \\
The sport is & table tennis & ice hockey \\
The measure is & heart rate & blood pressure \\
The role is & math teacher & history student \\
The item is & train ticket & hotel room \\
The event is & forest fire & ocean wave \\
The structure is & steel bridge & stone tower \\
The animal is & polar bear & sea turtle \\
The role is & software engineer & product manager \\
The object is & golden ring & silver coin \\
The instrument is & electric guitar & grand piano \\
The transport is & city bus & cargo train \\
The drink is & orange juice & green tea \\
The clothing is & winter jacket & summer dress \\
The service is & cloud storage & mobile network \\
The publication is & science book & travel guide \\
The landscape is & rose garden & pine forest \\
The building is & city library & village school \\
The communication is & email message & phone call \\
The schedule has & morning meeting & evening class \\
The artwork is & oil painting & pencil drawing \\
The finance term is & bank loan & credit score \\
The condition is & hand surgery & knee injury \\
The house part is & front door & back window \\
The media format is & news article & radio show \\
\bottomrule
\end{tabular}
\end{table}

\begin{table}[!htbp]
\centering
\small
\caption{
Complete three-mask prompt collection (20 prompts).
Each row instantiates the template:
\emph{The answer is either \{Alternative A\} or \{Alternative B\}:}
\texttt{[MASK][MASK][MASK]}.
}
\vspace{-8pt}

\label{tab:toy_3}
\setlength{\tabcolsep}{4pt}
\renewcommand{\arraystretch}{1.08}
\begin{tabular}{
    @{}
    >{\raggedright\arraybackslash}p{0.42\textwidth}
    >{\raggedright\arraybackslash}p{0.42\textwidth}
    @{}
}
\toprule
Alternative A & Alternative B \\
\midrule
red apple pie & green bean soup \\
silver sports car & black pickup truck \\
summer music festival & winter sports tournament \\
mountain bike trail & coastal hiking path \\
modern art museum & ancient history archive \\
solar energy research & marine biology laboratory \\
classical piano concert & modern dance performance \\
heart surgery team & cancer research center \\
primary school teacher & college football coach \\
mountain train journey & coastal bus route \\
tropical rain forest & desert sand storm \\
global sales report & local market survey \\
wooden kitchen cabinet & metal office desk \\
mobile phone charger & laptop power adapter \\
heavy winter snow & strong summer rain \\
senior software engineer & junior product designer \\
professional tennis player & college basketball coach \\
lunar research station & solar power satellite \\
spicy chicken curry & sweet apple pastry \\
urban public park & rural community center \\
\bottomrule
\end{tabular}
\end{table}

\begin{table}[!htbp]
\centering
\small
\caption{
Complete four-mask prompt collection (20 prompts).
Each row instantiates the template:
\emph{The answer is either \{Alternative A\} or \{Alternative B\}:}
\texttt{[MASK][MASK][MASK][MASK]}.
}
\vspace{-8pt}
\label{tab:toy_4}
\setlength{\tabcolsep}{4pt}
\renewcommand{\arraystretch}{1.08}
\begin{tabular}{
    @{}
    >{\raggedright\arraybackslash}p{0.42\textwidth}
    >{\raggedright\arraybackslash}p{0.42\textwidth}
    @{}
}
\toprule
Alternative A & Alternative B \\
\midrule
New York subway station & San Diego beach hotel \\
black leather office chair & white wooden dining table \\
summer music festival ticket & winter sports tournament pass \\
mountain bike repair shop & coastal hiking guide office \\
modern science fiction magazine & classic detective mystery novel \\
international airport security checkpoint & local railway ticket office \\
fresh fruit market stall & used book store counter \\
public health research center & private medical training school \\
mountain railway ticket office & coastal airport security gate \\
fresh vegetable soup recipe & warm chocolate cake recipe \\
elementary school science teacher & university history department chair \\
national football league game & local basketball team practice \\
tropical rain forest reserve & desert sand storm warning \\
global technology research company & local community health organization \\
heavy winter snow storm & strong summer rain shower \\
senior software engineering manager & junior product design specialist \\
international lunar research station & private orbital tourism company \\
spicy chicken curry recipe & sweet apple pastry recipe \\
coastal wildlife protection program & urban water conservation project \\
secure cloud storage service & public mobile network provider \\
\bottomrule
\end{tabular}
\end{table}

\end{document}